%% file: main.tex
\pdfoutput=1

\documentclass[11pt]{article}

\usepackage[final]{coling}

\usepackage{times}
\usepackage{listings}

\usepackage{latexsym}
\usepackage{amsfonts}
\usepackage{float}

\usepackage{fancyhdr}

\usepackage{bm}
\usepackage[T1]{fontenc}

\usepackage[utf8]{inputenc}
\usepackage{microtype}

\usepackage{inconsolata}

\usepackage{graphicx}
\usepackage[most]{tcolorbox}

\newcommand{\eg}{\textit{e.\,g.}}
\begin{document}
\title{Large Language Models are Good Multi-lingual Learners : \\ When LLMs Meet Cross-lingual Prompts}

\author{
  \textbf{Teng Wang\textsuperscript{1}\thanks{In Proceedings of the 31st International Conference on Computational
    Linguistics, pages 4442–4456, Abu Dhabi, UAE. Association for Computational Linguistics. Copyright 2025
by the author(s).}},
  \textbf{Zhenqi He\textsuperscript{1}},
  \textbf{Wing-Yin Yu\textsuperscript{2}},
  \textbf{Xiaojin Fu\textsuperscript{2}},
  \textbf{Xiongwei Han\textsuperscript{3}}
\\
  \textsuperscript{1}Department of Mathematics, The University of Hong Kong, Hong Kong SAR, China\\
  \textsuperscript{2}Noah's Ark Lab, Huawei, Hong Kong SAR, China\\
  \textsuperscript{3}Noah's Ark Lab, Huawei, Shenzhen, China\\
  \small{
    \textbf{Correspondence:} 
    \{wt0318, zhenqi\_he\}@connect.hku.hk, \{rocket.YuWingYin, 
fuxiaojin, hanxiongwei\}@huawei.com, 
    %
  }
}

\maketitle
\input{Secs/0_abstract}
\input{Secs/1_intro}

\input{Secs/2_relatedWork}

\input{Secs/3_method}

\input{Secs/4_exp}
\input{Secs/5_conclusion}

\input{Secs/6_limitation}

\bibliography{ref}
\newpage

\input{Secs/7_appendix}

\end{document}

%% file: Secs/0_abstract.tex
\begin{abstract}
With the advent of Large Language Models (LLMs), generating rule-based data for real-world applications has become more accessible. 
Due to the inherent ambiguity of natural language and the complexity of rule sets, especially in long contexts, LLMs often struggle to follow all specified rules, frequently omitting at least one. 
To enhance the reasoning and understanding of LLMs on long and complex contexts, we propose a novel prompting strategy \textbf{M}ulti-\textbf{L}ingual \textbf{Prompt}, namely \textbf{MLPrompt}, which automatically translates the error-prone rule that an LLM struggles to follow into another language, thus drawing greater attention to it.
Experimental results on public datasets across various tasks have shown MLPrompt can outperform state-of-the-art prompting methods such as Chain of Thought, Tree of Thought, and Self-Consistency. 
%
%
%
Additionally, we introduce a framework integrating MLPrompt with an auto-checking mechanism for structured data generation, with a specific case study in text-to-MIP instances. Further, we extend the proposed framework for text-to-SQL to demonstrate its generation ability towards structured data synthesis.
%
\end{abstract}

%% file: Secs/1_intro.tex
\section{Introduction}
Mixed Integer Programming (MIP) is a significant part of Operations Research (OR), which aims to solve optimization problems where some decision variables are constrained to be integers. It has been widely applied in industrial fields including logistics~\cite{hulagu2020mixed}, scheduling~\cite{mip_in_scheduling}, and supply chain management~\cite{mip_in_supplyChain}. Nowadays, benefiting from the development of Large Language Models (LLMs), automatically modeling complex and practical OR problems in plain text description to mathematical optimization formulas is no longer an impossible mission~\cite{xiao2024chainofexperts, cot, yao2023treethoughtsdeliberateproblem}. CoE~\cite{xiao2024chainofexperts} proposes a multi-agent system leveraging LLMs to model and program complex operations research problems and constructs a new dataset named ComplexOR involving intricate constraints, domain-specific terminology, and multi-step reasoning for various domains \eg supply chain, scheduling, and logistics.
%
%
Although the ComplexOR dataset provides foundational model metadata including \textit{‘set’}, \textit{‘parameter’}, \textit{‘hyper-parameter’}, \textit{‘variable’}, \textit{‘objective function’}, and \textit{‘constraint’}, concrete values of \textit{‘set’} and \textit{‘parameter’} are missing to generate MIP instances, causing limited applicability in developing autonomous MIP solvers for real-world optimization tasks.
%

\begin{figure}[t!]
\centering
\includegraphics[width=0.4\textwidth]{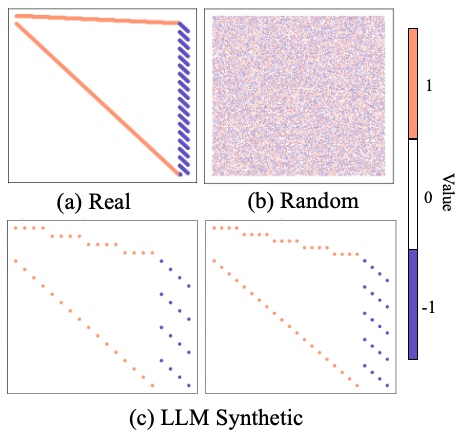}
\caption{Semantic illustrations of the potential of LLMs in generating data following real distributions. 
(a) Distribution of Constraint Coefficients in Real-World Factory Location Problems~\cite{cornuejols1977exceptional}.
(b) Distribution of Constraint Coefficients generated by random simulation.
(c) Distribution of Constraint Coefficients generated by GPT-$4$~\cite{achiam2023gpt} with different hyper-parameters.
For each case: (demand points, candidate locations) = (4, 4), (5, 4).
} 
\label{fig:dist}
\end{figure}

With the success of data synthesis in various domains~\cite{yang2024synthesizingtexttosqldataweak, zhang1,zhang3, li2023syntheticdatagenerationlarge, ArtiFusion, zhang2}, LLMs have shown great capabilities in synthesizing realistic data in multiple modalities while the potential of LLMs in generating MIP instances data has not been fully explored. By comparing the synthetic distributions generated by GPT-4~\cite{achiam2023gpt} with the real distribution of Constraint Coefficients of MIP instances and randomly generated distributions for Factory Location Problem~\cite{cornuejols1977exceptional}, shown in Fig.~\ref{fig:dist}, LLMs have demonstrated its superiority in MIP instances generating compared with random simulation.

\begin{figure}[t!]
\centering
\includegraphics[width=0.45\textwidth]{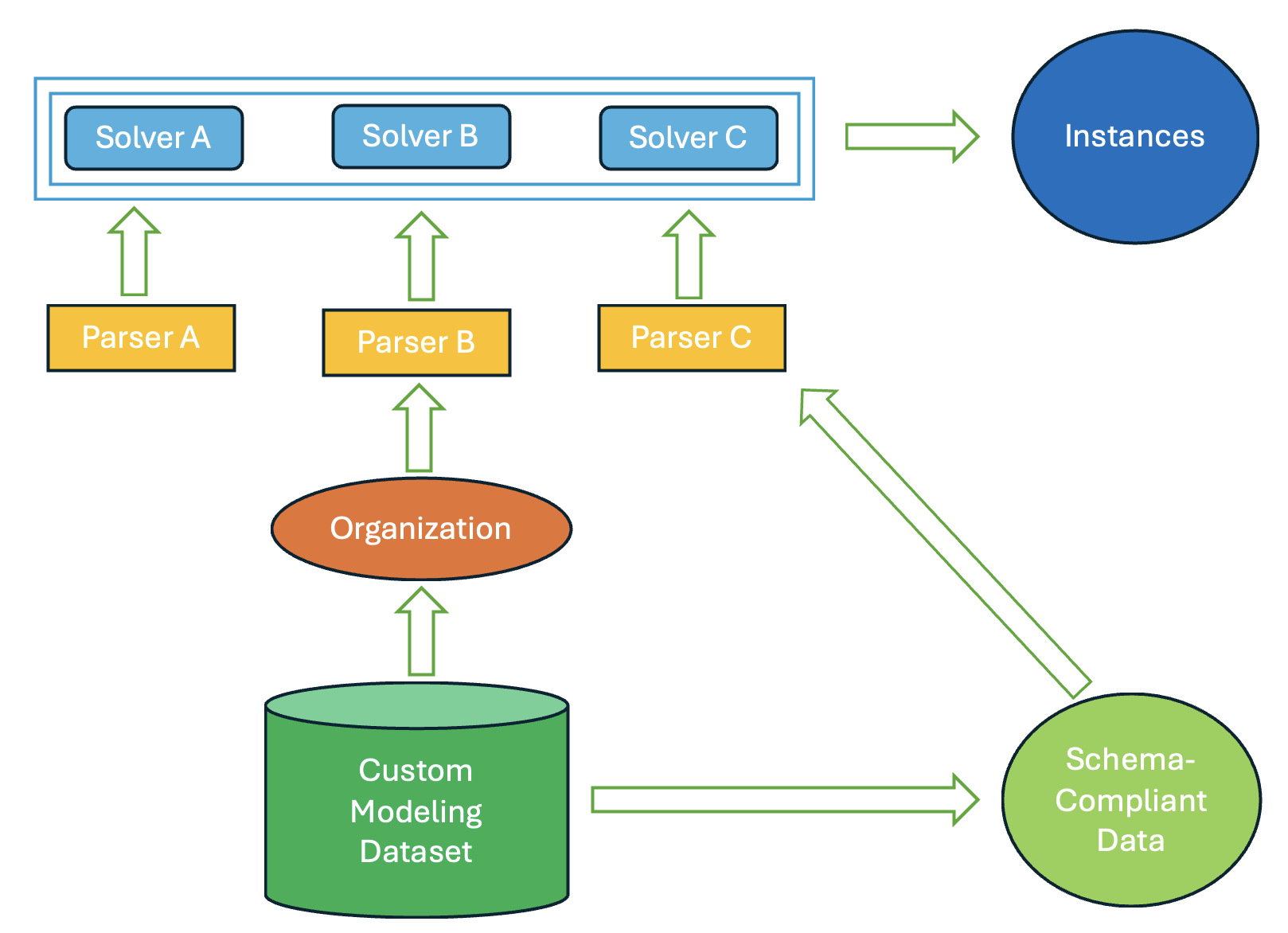}
\caption{Illustration of the MIP instance generation process using key modeling components, including sets, parameters, variables, constraints, and objective functions. Only when compliant with the parser rules can the parser generate the modeling code, and the modeling tools produce instances.
}

\label{fig:generating_pipeline}
\end{figure}

Fig.~\ref{fig:generating_pipeline} demonstrates the autonomous MIP instance generation pipeline, incorporating modeling information such as sets, parameters, variables, constraints, and objective functions across multiple solvers. 
%
Commercial MIP solvers such as Gurobi~\cite{achterberg2019s}, OptVerse~\cite{li2024machine}, and CPLEX~\cite{bliek1u2014solving} usually have its unique data representation with different parser rules for MIP instances, requiring various intricate rules in natural language to condition the generation of data that can be imported by the solvers for successful modeling. While direct handcrafted rules may lead to long and ambiguous contexts, LLMs have a high risk of neglecting certain rules to generate unsatisfactory instances given the complicated contexts.

%

%

To facilitate
the autonomous generation of MIP instances within industrial pipelines, we propose a novel \textbf{M}ulti-\textbf{L}ingual \textbf{P}rompt algorithm, namely \textbf{MLPrompt}, specifically designed for structural data synthesis and adaptability for various solvers with different formats.
Additionally, we propose a framework incorporating MLPrompt with an auto-checking mechanism to enable iterative prompt updates to ensure compliance with input constraints.
%
%

\begin{figure}[t!]
\centering
\includegraphics[width=0.45\textwidth]{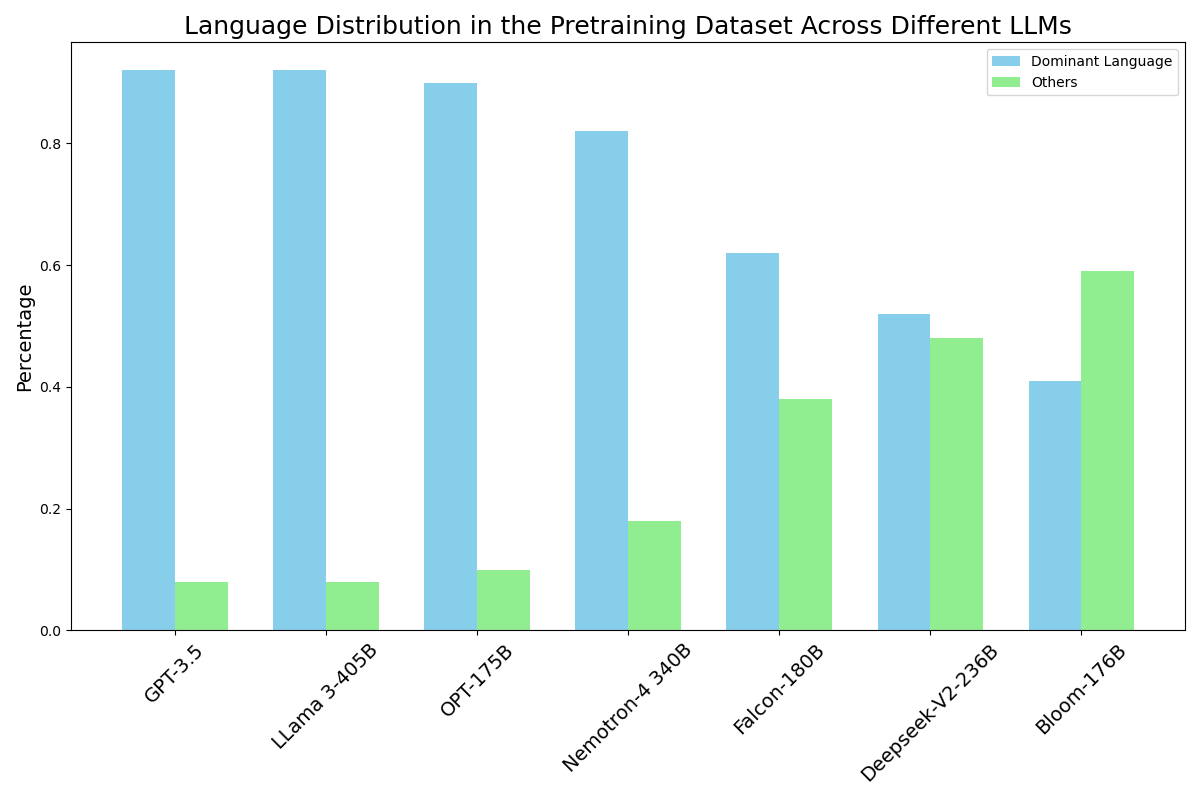}
\caption{The dominant language refers to the majority language in the pretraining dataset, while "others" refer to all remaining languages.
 This figure demonstrates the language imbalance phenomenon in the pretraining data of GPT-$3.5$~\cite{brown2020language}, Llama-$3$~\cite{llama3}, Deepseek-V$2$~\cite{deepseek}, Bloom~\cite{bloom}, Nemotron-4~\cite{adler2024nemotron}, OPT~\cite{zhang2022opt}, and Falcon~\cite{almazrouei2023falcon}.}
\label{fig:language_distribution}
\end{figure}

The design of MLPrompt is motivated by the following observations. The trilingual parallel language processing experiments~\cite{Pathak_Vulchanova_Pathak_Mishra_2024} have shown that more dominant languages receive greater cognitive focus, which facilitates faster response times and reduces the mental load when processing in non-dominant languages for polyglots. These findings are derived from mouse-tracking experiments~\cite{Pathak_Vulchanova_Pathak_Mishra_2024} that observed participants’ language processing while listening to words in different languages and selecting corresponding images.
%
Similarly, LLMs, such as ChatGPT~\cite{achiam2023gpt}, function as polyglots, supporting over 80 languages. The scale of pretraining data varies across languages, as depicted in Fig.~\ref{fig:language_distribution}, where dominant languages frequently appear in the pretraining data, while non-dominant languages are comparatively underrepresented
%
%
~\cite{achiam2023gpt}. Hence, deriving from phenomena in human polyglots where the existence of non-dominant languages helps the process of dominant languages, MLPrompt is proposed to leverage the non-dominant languages of LLMs to strengthen the understanding on dominant languages.



%

Existing LLM-based reasoning approaches, such as Chain-of-Thought (CoT)~\cite{cot}, Tree-of-Thoughts (ToT)~\cite{yao2023treethoughtsdeliberateproblem}, Self-Consistency (SC)~\cite{wang2023selfconsistencyimproveschainthought}, and ReAct~\cite{yao2023react}, primarily focus on enhancing reasoning through iterative steps. These approaches aim to improve the quality of generated outputs by verifying the correctness of intermediate steps and selecting the best path forward. However, when generating structured outputs like JSON that must adhere to given constraints, it poses challenges to break the task into smaller and independent parts due to the interconnections within each part. 
%
%
In contrast, MLPrompt tackles structured data synthesis with introducing non-dominant languages to raise LLMs' attention in error-prone constraints reduce inference time without requiring multiple inference steps, all while preserving the interconnected relevance of the constraints.

In this paper, we propose a general MIP instances synthesis pipeline with a novel prompting method. Our contributions are three-fold: (1) We introduce \textbf{MLPrompt}, a simple yet effective multiple-lingual prompting strategy, for enhancing LLM reasoning, inspired by the capability of cross-lingual understanding in polyglots.
%
%
(2) We make the first attempt at a LLM for MIP instance generation which serves as a bridge to connect existing research-based datasets with industrial needs, and it is easy to be extended into a general pipeline for structured data synthesis.
%
(3) Extensive experiments on the ComplexOR dataset demonstrate the superiority of our prompt strategy compared to existing prompting strategies for MIP instance generation. Moreover, additional demonstration on the text2SQL\cite{yu2018spider,cao2024spider2} task highlights the broader applicability of our framework in other structured data generation tasks.

%% file: Secs/2_relatedWork.tex
\section{Related Work}
\subsection{MIP Instance Generation}

MIP instance generation plays a crucial role in the development of commercial MILP solvers~\cite{mip_solver}. Traditional mathematical-formulation-based methods for MIP instance generation include TSP~\cite{mip_instance_generation_mathMethod}, structure-based instance generation~\cite{bixby2000mip, applegate2006traveling}, mixed-integer knapsack~\cite{mip_knapsack}, and set covering~\cite{mip_setCovering, mip_setcovering2}. G2MILP~\cite{geng2024deepinstancegenerativeframework} introduces the first learning-based MILP instance generation framework, representing MILP problems as bipartite graphs and using a masked variational autoencoder (VAE)~\cite{VAE} to iteratively generate new instances. ACM-MILP~\cite{guo2024acmmilp} presents an adaptive and structure-preserving approach by using a community detection algorithm to group strongly related constraints for collective modification. MILPGen~\cite{mipGen2024} simplifies bipartite graph representations of MILP instances into tree-like structures and uses graph convolutional networks (GCNs) to calculate node embeddings, allowing for optimal node pair merging. This enables the creation of larger, more complex instances while maintaining the original structural characteristics. Building upon these approaches, our method generates general MILP instances for a variety of applications, such as scheduling, logistics, and product management, through the use of input text descriptions. Specifically, we leverage the modeling information from the ComplexOR dataset to produce MIP instances, aligning with the broader goal of addressing real-world industrial needs.



\subsection{Prompt Engineering for LLMs}Prompts are gradient-free strategy to strengthen LLMs' reasoning for complex tasks. The Chain-of-Thought (CoT)~\cite{cot} decouples complex reasoning tasks into intermediate reasoning steps. Auto-CoT~\cite{zhang2023autocot} automate CoT by by encouraging LLMs to \textit{think step by step} to reduce manual operations in prompting. Self-Consistency (SC)~\cite{wang2023selfconsistencyimproveschainthought} introduces a novel decoding strategy to enhance the reasoning capabilities of LLMs when using CoT prompting. Instead of relying on a single reasoning path derived from greedy decoding, it samples multiple diverse reasoning paths and determines the final answer by marginalizing over these paths to select the most consistent one. Tree-of-thought (ToT)~\cite{yao2023treethoughtsdeliberateproblem} and Graph-of-thought (GoT)~\cite{graphOfThought} enhance the reasoning capabilities of LLMs through structured prompting schemes that go beyond traditional linear approaches like CoT.

\subsection{Multilingual LLMs}
With the success of English-center LLMs across various NLP tasks such as Question Answering~\cite{NarrativeQA} and Summarization~\cite{summarization}, increasing attention has been drawn to multilingual LLMs due to globalization. A multilingual LLM possesses the ability to process and produce content in multiple languages simultaneously. Existing approaches for Multilingual LLMs are mainly split into two ways - (1) Parameter-tuning the LLMs with multilingual data~\cite{du2022glm, mendonca-etal-2023-towards} (2) Parameter-frozen with prompting~\cite{hoang-etal-2024-fly}. In this paper, we focus on the analysis of increasing the understanding and reasoning capabilities of LLMs trained with multilingual data such as GPT~\cite{achiam2023gpt} with proposed MLPrompt.

%% file: Secs/3_method.tex



\section{Methodology}
\subsection{Problem Statement}
\label{problem}

\begin{figure}[t!]
\centering
\includegraphics[width=0.45\textwidth]{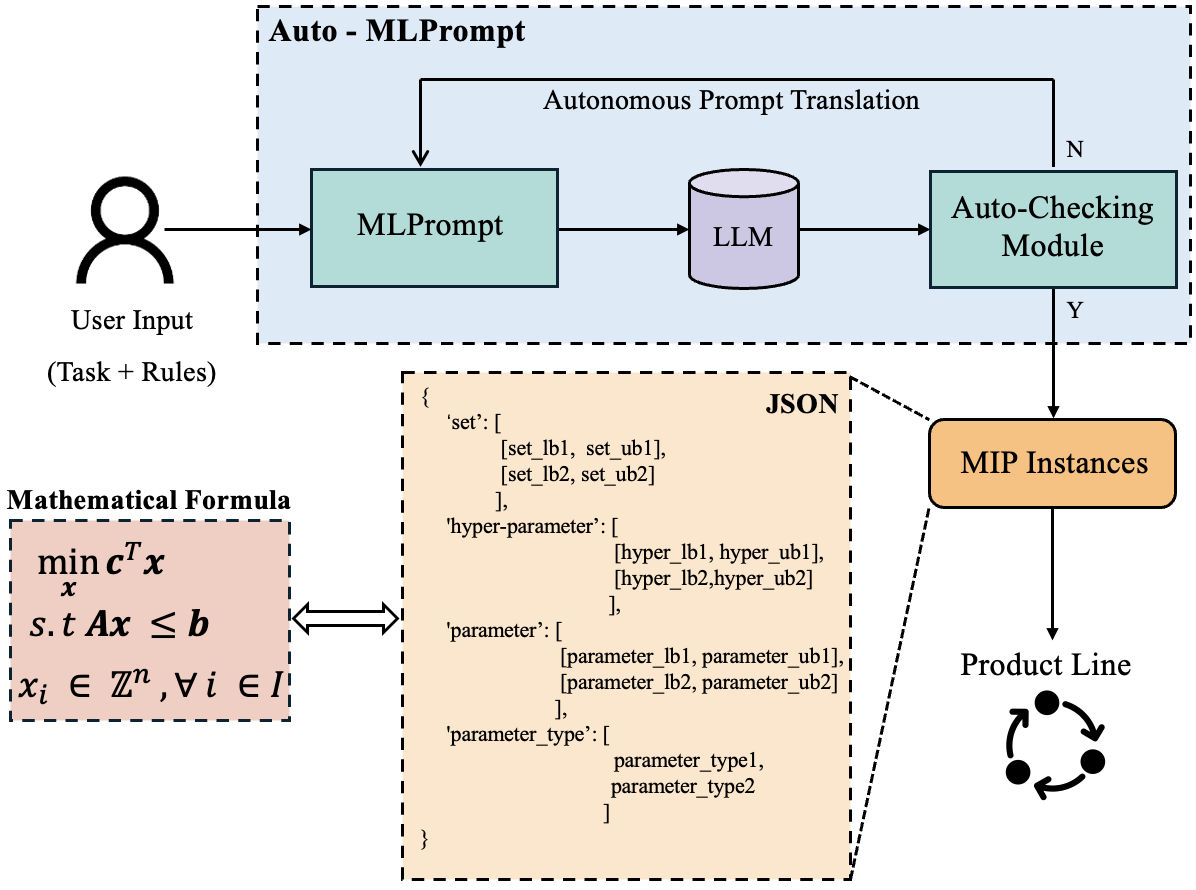}
\caption{The workflow of the proposed framework for structured data generation. The process includes prompt construction with predefined rules, data generation by the LLM, evaluation of compliance with the rules, and iterative refinement of the prompt by translating rules into other languages if necessary.
} 
\label{fig:XGen}
\end{figure}

An Mixed-Linear-Integer-Programming (MLIP) problem can be formulated as follows:

\begin{equation}
    \centering
    \begin{split}
        &\min_{\bm{x}} \bm{c}^T \bm{x} \\
        & \text{s.t. } \bm{A} \bm{x} \leq \bm{b}, \\
        & \bm{x}_i \in \mathbb{Z}^n \text{ for } \forall i \in I,\\
    \end{split}
\end{equation}

where $\bm{c} \in \mathbb{R}^n$ is the vector of objective coefficients, $\bm{A} \in \mathbb{R}^{m \times n}$ is the matrix of constraint coefficients, $\bm{b} \in \mathbb{R}^m$ is the vector of constraint bounds, and $\bm{x} \in \mathbb{R}^n$ represents the decision variable. The index set $I$ indicates the decision variables $\bm{x}_i$ constrained to be integers. 

Here, $\bm{A}$, $\bm{b}$, and $\bm{c}$ are parameters to be generated, and their dimensions $n, m$ will be defined in the given modeling problem. To simplify the generation and avoid unexpected long outputs, we formulate the MLIP instances into JSON data to record the data types, along with the lower and upper bounds for sets, parameters, and hyper-parameters as shown in Fig.~\ref{fig:XGen}. Then a random process generates values within the generated bounds to be fed into the solver's parser to construct the MLIP instance. By that, we successfully convert the MLIP instance generation problem into rule-based structured data generation, and the core is to employ the LLM to produce the corresponding JSON.




\begin{figure}[t!]
\centering
\includegraphics[width=0.45\textwidth]{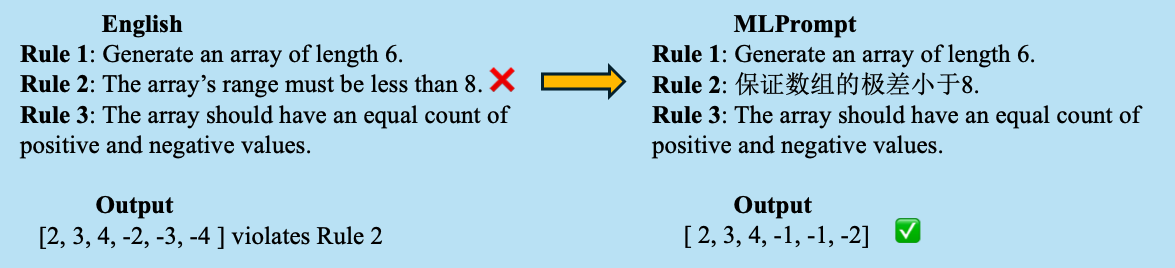}
\caption{A demo of how MLPrompt builds prompts.}

\label{fig:MLPrompt}
\end{figure}

\subsection{MLPrompt}
Generally, user-specific parsers are used in industries and academic teams for mathematical modeling and MIP instance generation~\cite{xing2024towards, MIP2solver}, leading to complex and interdependent rules to condition the generation of desired structured data.
%
To address the complex constraints in natural language, we propose \textbf{MLPrompt} to leverage the multilingual capabilities of LLMs to draw LLMs attention to error-prone rules by translating any rule that is not followed into a non-dominant language of LLMs and hence to improve the quality of the generated data.

\textbf{MLPrompt}
Inspired by the phenomenon in human polyglots where the existence of non-dominant languages helps the process of dominant languages~\cite{Pathak_Vulchanova_Pathak_Mishra_2024}, we propose the \textbf{MLPrompt} to translate the error-prone rule into a non-dominant language of LLM to strengthen the understanding and reasoning of LLM in given complex contexts. A demonstration of this approach is presented in Fig.~\ref{fig:MLPrompt}.
%
%

Here, we define the dominant language as the most frequently occurring language within the pretraining dataset, while all other languages, except the dominant one, are referred to as "others.". Prior work on Falcon~\cite{almazrouei2023falcon} highlights the challenge that arises when incorporating a substantial amount of multilingual data (e.g., exceeding 10\%) in language models—leading to a decline in performance on tasks that are more aligned with the dominant language. Consequently, many state-of-the-art LLMs prioritize training on large volumes of dominant language data, further exacerbating the imbalance across languages.
Fig.~\ref{fig:language_distribution} demonstrates that while LLMs are designed with multilingual capabilities in mind, there is a prevalent tendency to disproportionately prioritize dominant language exposure, primarily to optimize performance on tasks aligned with the dominant language.


\textbf{Auto-MLPrompt Strategy} 
As introduced in Sec.~\ref{problem}, we formulate the MIP instance generation as a conditioned structured data synthesis constrained by natural language rules. It is relatively straightforward to validate whether JSON data adheres to the given constraints and to identify the specific rule or rules that have been violated. With such property in the rule-based JSON generation, we design an autonomous MIP generation pipeline with the proposed MLPrompt strategy to automatically detect the violated rule and translate it into a non-dominant language. The generation flow can be summarized as follows: 
%
%
Initially, we incorporate the predefined rules and input modeling information directly to generate the initial prompts.
Then, the LLM generates structured data, like JSON, based on the initial prompt, and an evaluation function assesses whether the generated data complies with the given rules. If the error or mis-generation is detected, the corresponding rule will be translated into a non-dominant language to update the prompt, and the data generation process will be repeated.

%% file: Secs/4_exp.tex
\section{Experiments}
\subsection{Dataset}
\label{dataset}
ComplexOR~\cite{xiao2024chainofexperts} contains $60$ complex operations research problems with natural language description and corresponding mathematical formula. For each problem, it contains comprehensive information (an example is shown in Appendix~\ref{sec:appendix:complexor}) for constructing models, including problem backgrounds, sets, parameters, hyper-parameters, variables, objective functions, and constraint functions. To generate the data following given conditions, we convert this extensive context into a set of rules for the LLMs to follow, requiring the output JSON to include the data type, lower bound, and upper bound for each set, parameter, and hyper-parameter. Subsequently, a random process is simulated to generate values for these sets and parameters. The solver's parser then reads the model-building information, loads the generated random data, and constructs the MIP instance accordingly.

\subsection{Experimental Settings}
As described in Section \ref{dataset}, we use the ComplexOR~\cite{xiao2024chainofexperts} dataset (an example of binpacking problem is shown in Appendix ~\ref{sec:appendix:complexor}) for our experiments and utilize an LLM to generate a JSON file constrained by input rules. The rules are outlined in Appendix \ref{sec:appendix:text2mip}, and we evaluate the model's performance based on rules 4, 7, and 8. A detailed analysis of the model's compliance with these rules is provided in Appendix \ref{sec:appendix:analysis_text2MIP}. We also implement an evaluation function to check whether the generated JSON complies with each rule, compute the accuracy for each rule, and calculate the final score by averaging the accuracies of the three rules.


We evaluate our proposed prompting strategy, MLPrompt, on various LLMs of different sizes, categorizing them into three groups based on the number of parameters. (1) \textbf{Small-scale LLMs}: LLMs with fewer than $10$ billion parameters are classified as small. We use open-source models such as Mistral-7B~\cite{mistral7B}, Llama-3-8B~\cite{llama3}, Gemma-2-9B~\cite{gemma2}, Qwen2-7B~\cite{qwen2}, Llama-3.1-8B~\cite{llama3}, Qwen1.5-7B~\cite{qwen15} and Openchat-3.5-7B~\cite{openchat}. 
(2) \textbf{Medium-scale LLMs}: LLMs with parameter sizes between $50$ billion and $200$ billion are considered medium-sized. These include GPT-3.5~\cite{brown2020language}, Mixtral-8$\times$7B~\cite{mistral56B}, Llama-3.1-70B~\cite{llama3}, Deepseek-67B~\cite{deepseek}, Llama-3-70B~\cite{llama3}, Qwen2-72B~\cite{qwen2}, WizardLM-8$\times$22B~\cite{wizardlm} and Mixtral-8$\times$22B~\cite{mistral56B}. 
(3) \textbf{Large-scale LLMs}: We define LLMs with more than $200$ billion parameters as large models. For this category, we focus on the GPT-4 series and conduct experiments using GPT-4o, GPT-4o mini, and GPT-4-Turbo~\cite{achiam2023gpt}.

For each model scale, we compare MLPrompt with CoT~\cite{cot}, ToT~\cite{yao2023treethoughtsdeliberateproblem} and SC~\cite{wang2023selfconsistencyimproveschainthought}

\subsection{Small-scale LLMs}
\label{sec:exp:small_llms}
\input{Tabs/small_model}

In our experiments, the input contexts are long and comprehensive, requiring LLMs to have a strong capability of understanding and reasoning of long contexts, while small models often fail to comprehend our intentions and fail to generate the correct JSON format by input constraints. 
%
Hence, for experiments on the small-scale model, we only consider the success rate of generating the correct JSON format by LLMs, without assessing rule compliance. 


Natural language, unlike formal logic, lacks clear evaluative properties, making it difficult to assess the accuracy of generated outputs. This limitation poses a challenge for methods like CoT, ToT, and SC, which struggle to handle this ambiguity effectively. Additionally, the interconnected nature of rules in structured outputs, such as JSON, makes it difficult to verify each step in isolation—correctness in one part does not necessarily guarantee overall correctness when the parts are combined. To effectively utilize SOTA prompting methods, in the following experiments, CoT, ToT, and SC are applied without evaluating intermediate results. Detailed explanations are provided in Appendix \ref{sec:appendix:SOTA_prompt_methods}.


The performance, as shown in Table \ref{tab:small_model_inverted}, indicates that these small LLMs struggle to generate the desired JSON format and improve the quality of the generated data under prompting methods such as CoT, ToT, and SC. Due to their limitations in handling complex tasks, we will not use these small LLMs in further experiments.

\subsection{Medium-scale LLMs}
\input{Tabs/medium_model}

Medium LLMs have the capability to comprehend our requirements and generate correctly formatted JSON. We evaluate these medium LLMs to determine whether the generated data adhere to the rules we have defined. The final results, as shown in Table \ref{tab:medium_model}, demonstrate that MLPrompt effectively improves the quality of generating complex data by translating a single rule from one language to another. This approach is more efficient compared to methods like ToT and SC, which involve multiple steps to obtain intermediate results, then combine and infer from these results to reach the final output.

The poor performance of CoT, ToT, and SC, as shown in Table \ref{tab:medium_model}, can be attributed to our inability to evaluate the intermediate results required by these methods in generating complex data task. As discussed in Section \ref{sec:exp:small_llms}, these medium LLMs can be considered as weak learners, and their intermediate outputs often contain errors. As a result, combining multiple weak learners not only fails to improve the quality of the final results but may even degrade them. In contrast, similar to bagging, only when combining the results of strong learners can performance improvements be expected \cite{bhavan2019bagged}.

Furthermore, as shown in Table \ref{tab:medium_model}, adding the missing rule in another language alongside the existing one not only fails to outperform the replacement method but, in some cases, performs worse than the baseline. This approach also appears influenced by the re-adding prompt method, which undermines the purpose of the ablation experiment. Therefore, in the subsequent experiments, we will focus solely on the MLPrompt strategy using the replacement approach.

\subsection{Large-scale Models}
\input{Tabs/large_model}

In this section, we evaluate the performance of large-scale models, specifically from the GPT-4 series, across various prompting strategies.
GPT-4o served as the base model, and the results, as shown in Table \ref{tab:large_model}, demonstrate several key findings. 
The baseline performance in a zero-shot setting is reasonable, and improvements are achieved using CoT, ToT, and SC prompting methods, which showcase their ability to refine model performance by structuring reasoning steps.

However, the most significant gains are observed with the MLPrompt strategy, which replaces key rules with alternative languages such as Mandarin, Thai, and Korean. Across all model variations, MLPrompt consistently yielded the highest performance, with Mandarin replacement proving particularly effective. This strategy outperformed traditional multi-step prompting methods like CoT, ToT, and SC, not only improving accuracy but also enhancing the efficiency of the models by reducing inference time and improving the overall quality of the generated data. As such, MLPrompt demonstrates its superiority in handling complex data generation tasks in large-scale models.

\begin{figure}[H]
\centering
\includegraphics[width=0.5\textwidth]{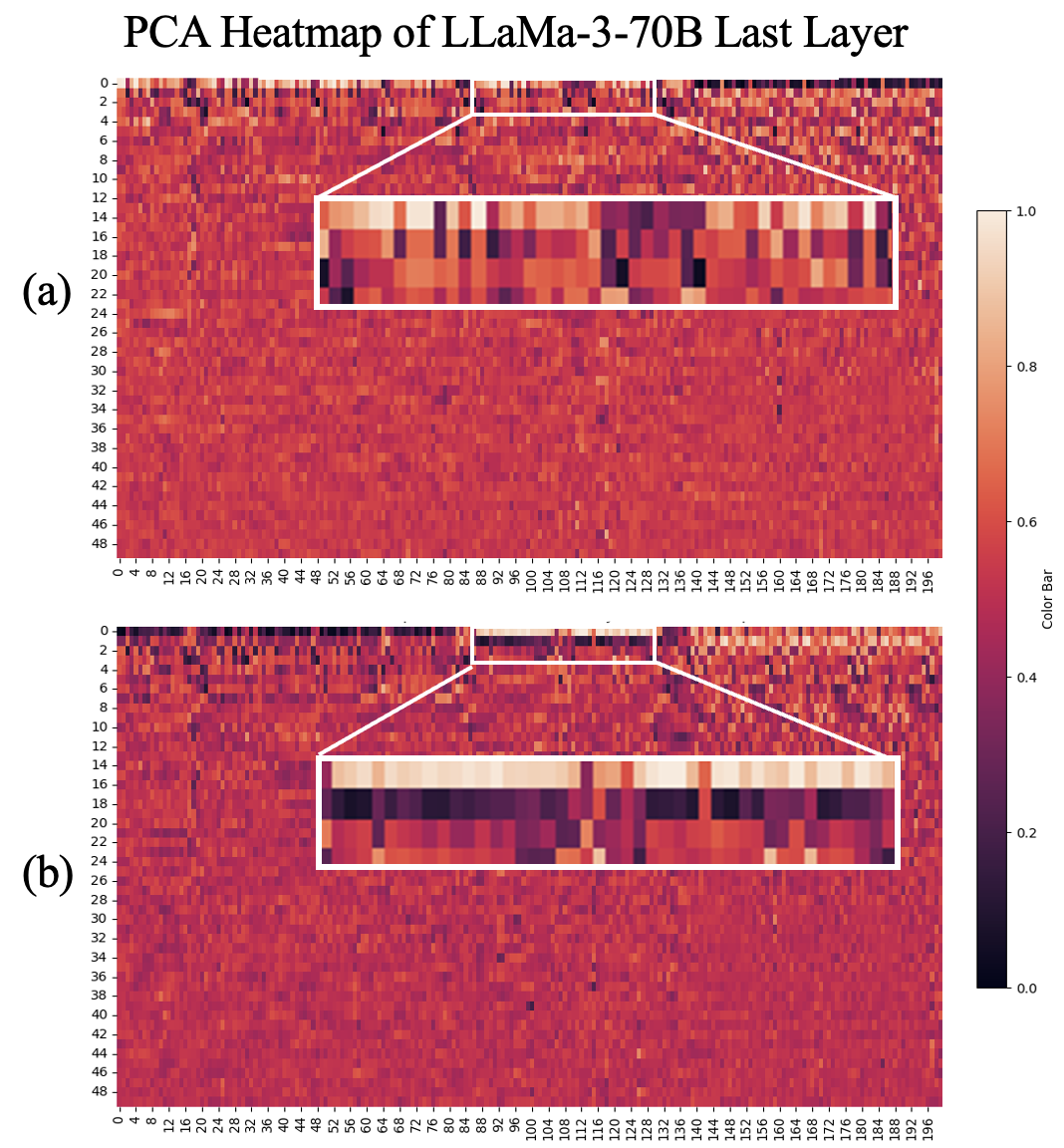}
\caption{Heatmap of PCA-transformed final layer of LLaMA-3-70B with different prompts. (a) With original single-lingual Prompt. (b) Multi-lingual Prompts with the rule 8 in maindarin (Ours). The X-axis represents input tokens, and the Y-axis shows the 50 PCA components.  
}
\label{fig:combinedAttention}
\end{figure}

\subsection{Attention Verification}
To verify whether MLPrompt increases the attention of LLMs of error-prone rules, we visualize the attention map of the final layer of the open-source LLaMA-3-70B~\cite{llama3} when inputting the English prompt and our MLPrompt with rule 8 being translated into Mandarin. The attention map is firstly downsampled from $8192$ to $50$ for better representation using PCA. Fig.~\ref{fig:combinedAttention} shows the downsampled attention map, and the corresponding error-prone rule 8 positioned between 84 to 128 
has been zoomed in for clear representations. The first few principal components of our proposed prompts predominantly focus on the translated Rule 8 while the original English prompts do not exhibit a similar concentration on any specific rule. The attention map visualization from the final layer of the LLM illustrates the efficacy of our proposed MLPrompt, which effectively directs the LLM’s attention towards error-prone rules by incorporating non-dominant languages.
%


\subsection{Text-to-SQL}
\label{sec:experiment:text2sql}
We further expand our MLPrompt in text-to-SQL tasks to validate the generalization ability.
Text-to-SQL is more challenging than text-to-JSON due to the difficulty in detecting rule violations in SQL, which in turn complicates the identification of error-prone rules.
%
However, it is straightforward to manually identify rule violations from the SQL results. To demonstrate the effectiveness of our approach, we present an example where MLPrompt outperforms other methods in text-to-SQL, and the rules for LLMs and the process by which these rules are derived are detailed in Appendix \ref{sec:appendix:text2sql}.

The task is: "Find the first name and age of students who have a pet," using the pets\_1 database from the Spider V1.0 dataset \cite{yu2018spider}. We generate the prompt for GPT-4 by combining the SQL schema with predefined rules. Manual analysis reveals that GPT-4 frequently violates rule 4 in this task. To address it, we translate rule 4 into Korean, Japanese, and Mandarin respectively using an online translation. 
%
We also present ablation studies examining repetitive error-prone rules. 
The performance, evaluated by the error rate for SQL execution based on 20 runs of the provided example, is shown in Tab.~\ref{tab:appendix_text2sql}, which shows that GPT-4 would fail to follow long and comprehensive rules. Even when the error-prone rules are repeated for emphasis, GPT-4 still fails. 
%
A potential strategy to mitigate this challenge is to split the rules within the prompt. However, isolating them is not feasible due to their interleaved nature, which could compromise the integrity of the prompt structure.
%
%
Our proposed MLPrompt, utilizing various non-dominant languages, consistently exhibits low error rates, demonstrating the superiority of MLPrompt in text-to-SQL tasks.

\input{Tabs/appendix_text2sql}

%% file: Tabs/small_model.tex
\begin{table*}[]

\centering
\resizebox{\linewidth}{!}{
\begin{tabular}{l|c|c|c|c|c|c|c}
\hline
Methods    & Mistral-7B & Llama-3-8B   & Llama-3.1-8B & Gemma-2-9B & Qwen1.5-7B & Qwen2-7B  & Openchat-3.5-7B\\
\hline
Zero-shot    &  0.211  &  0.268  &  0.016  &  0.302  &  \textbf{0.017}  &  0.100  &  0.250 \\
Few-shots    &  0.178  &  0.062  &  0.032  &  0.326  &  0.017  &  0.033  &  0.100 \\
CoT\cite{cot}          &  \textbf{0.350}  &  \textbf{0.300}  &   0.167  &  \textbf{0.717}  & 0.000  &  0.133  &  0.233\\
ToT\cite{yao2023treethoughtsdeliberateproblem}        &  0.300  &  0.033  &  \textbf{0.350}  &  0.033  &  0.000  &  0.233  &  \textbf{0.433}\\
SC\cite{wang2023selfconsistencyimproveschainthought}          &  0.267  &  0.067  &  0.333  &  0.050  &  0.000  &  \textbf{0.250}  &  0.100 \\
\hline
\end{tabular}
}
\caption{
The table presents the success rate of small-scale LLMs generating correctly formatted JSON under various prompting methods, without considering rule compliance.
}
\label{tab:small_model_inverted}

\end{table*}



%% file: Tabs/medium_model.tex

\begin{table*}[]
\footnotesize

\centering
\resizebox{\linewidth}{!}{
\begin{tabular}{l|l|l|l|l|l|l|l|l}
\hline
Methods    & GPT-$3.5$ & WizardLM-8$\times$22B  & Mixtral-8$\times$7B & Mixtral-8$\times$22B & Llama-$3.1$-$70$B & Llama-3-$70$B & Deepseek-$67$B &  Qwen$2$-$72$B \\
\hline
Baseline   &  0.436  &  0.250  &  0.133  &  0.239 &  0.378  &  0.461  &  0.083  &  0.383  \\

\hline

CoT\cite{cot}  &  0.156  &  0.389  &  0.128  &  0.261  &  0.428  &  0.150  &  \textbf{0.128}  &  0.583   \\
ToT\cite{yao2023treethoughtsdeliberateproblem}  &  0.294  &  0.339  &  0.011  &  0.128 &  0.167  &  0.083  &  0.061  &  0.606  \\
SC\cite{wang2023selfconsistencyimproveschainthought}  &  0.131  &  0.217  &  0.011  &  0.078 &  0.167  &  0.117  &  0.000  &  0.611   \\
Repeated Missed Rule  &  0.133  &  0.244  &  0.067  &  0.228 &  0.328  &  0.183  &  0.056  &  0.528   \\

\hline

MLPrompt + Mandarin  &  0.378  &   0.283 \textcolor{red}{$\uparrow$ 3.3\%}  &  0.039  &  0.239 &  0.472 \textcolor{red}{$\uparrow$ 9.4\%}  &  0.194  &  0.094 \textcolor{red}{$\uparrow$ 1.1\%}  &  \textbf{0.633} \textcolor{red}{$\uparrow$\textbf{ 25.0\%}}  \\
MLPrompt + Thai  &  0.383  &   0.272 \textcolor{red}{$\uparrow$ 2.2\%}  &  0.011  &  \textbf{0.306} \textcolor{red}{$\uparrow$ \textbf{6.7\%}} &  \textbf{0.500 \textcolor{red}{$\uparrow$ 12.2\%}}  &  0.372  &  0.044  &  0.500 \textcolor{red}{$\uparrow$ 11.7\%} \\
MLPrompt + Korean  &  0.372  &  0.350 \textcolor{red}{$\uparrow$ 10.0\%}  &  0.044  &  0.256 \textcolor{red}{$\uparrow$ 1.7\%} &  0.483 \textcolor{red}{$\uparrow$ 10.5\%}  &  0.233  &  0.050  &  0.572 \textcolor{red}{$\uparrow$ 18.9\%}   \\

\hline
MLPrompt $\leftrightarrow$ Mandarin  &  0.414  &  0.383 \textcolor{red}{$\uparrow$ 13.3\%}  &  \textbf{0.211} \textcolor{red}{$\uparrow$ \textbf{7.8\%}}  &  0.289 \textcolor{red}{$\uparrow$ 5.0\%} &  0.411 \textcolor{red}{$\uparrow$ 3.3\%}  &  0.483 \textcolor{red}{$\uparrow$ 2.2\%}  &  0.117 \textcolor{red}{$\uparrow$ 3.4\%}  &  0.422 \textcolor{red}{$\uparrow$ 3.9\%}  \\
MLPrompt $\leftrightarrow$ Thai      &  0.454 \textcolor{red}{$\uparrow$ 1.8\%}  &  0.294 \textcolor{red}{$\uparrow$ 4.4\%}  &  0.150 \textcolor{red}{$\uparrow$ 1.7\%}  &  0.294 \textcolor{red}{$\uparrow$ 5.5\%} &  0.394 \textcolor{red}{$\uparrow$ 1.6\%}  &  0.456  &  \textbf{0.128} \textcolor{red}{$\uparrow$ \textbf{4.5\%}}  &  0.428 \textcolor{red}{$\uparrow$ 4.5\%}   \\
MLPrompt $\leftrightarrow$ Korean    &  \textbf{0.591} \textcolor{red}{$\uparrow$ \textbf{15.5\%}}  &  \textbf{0.406} \textcolor{red}{$\uparrow$ \textbf{15.6\%}}  &  0.089  &  0.267 \textcolor{red}{$\uparrow$ 2.8\%} &  0.444 \textcolor{red}{$\uparrow$ 6.6\%}  &  \textbf{0.533} \textcolor{red}{$\uparrow$ \textbf{7.2\%}}  &  0.106 \textcolor{red}{$\uparrow$ 2.3\%}  &  0.339  \\
\hline

\end{tabular}}
\caption{
The table presents the accuracy of each medium-scale model across various settings. MLPrompt consistently enhances data quality. The "+" symbol denotes the addition of a new language rule to complement an unmet rule, while "$\leftrightarrow$" signifies the replacement of the original rule with an existing one.
} 
\label{tab:medium_model}

\end{table*}

%% file: Tabs/large_model.tex
\begin{table}[]
\footnotesize

\centering
\resizebox{\linewidth}{!}{
\begin{tabular}{l|l|l|l}
\hline
Methods    & GPT-$4o$  & GPT-$4o$ mini & GPT-$4$-Turbo \\
\hline
Baseline   &  $0.472$    &   $0.625$   &   $0.753$   \\
\hline
CoT\cite{cot}   &  $0.492$    &   -   &   -   \\
ToT\cite{yao2023treethoughtsdeliberateproblem}   &  $0.530$    &   -   &   -   \\
SC\cite{wang2023selfconsistencyimproveschainthought}   &  $0.619$    &   -   &   -   \\
Repeated Missed Rule  &  $0.536$    &   -   &   -   \\
\hline
MLPrompt $\leftrightarrow$ Mandarin &  $\textbf{0.844}$ \textcolor{red}{$\uparrow$\textbf{37.2\%}}    &   $\textbf{0.888}$  \textcolor{red}{$\uparrow$\textbf{26.3\%}}   &   $0.874$ \textcolor{red}{$\uparrow$12.1\%}  \\
MLPrompt $\leftrightarrow$ Thai     &  $0.813$ \textcolor{red}{$\uparrow$34.1\%}    &   $0.816$ \textcolor{red}{$\uparrow$19.1\%}   & $0.777$ \textcolor{red}{$\uparrow$2.4\%}    \\
MLPrompt $\leftrightarrow$ Korea    &  $0.796$ \textcolor{red}{$\uparrow$32.4\%}   &   $0.613$   &   $\textbf{0.892}$ \textcolor{red}{$\uparrow$\textbf{13.9}\%}          \\
\hline
\end{tabular}}
\caption{
Performance evaluation of large-scale LLMs, comparing GPT-4 series under different prompting strategies, including CoT, ToT, SC, and Repeated-missed-rule.
}
\label{tab:large_model}

\end{table}


%% file: Tabs/appendix_text2sql.tex
\begin{table}[]
    \centering
      
    \begin{tabular}{c|c}
        \hline
        Prompt combination & Error Rate (\%) \\
        \hline
        English & 0.35 \\
        English w Repetitive  Rule 4 & 0.50 \\
        \hline
        Rule 4 in Mandarin (Ours) & 0.10 \\
        Rule 4 in Japanese (Ours) & \textbf{0.00} \\
        Rule 4 in Korean (Ours)& \textbf{0.00} \\
        \hline
    \end{tabular}
      \caption{Error rates for different prompt configurations in the text-to-SQL task, based on 20 runs of the given sample.
}    \label{tab:appendix_text2sql}

\end{table}

%% file: Secs/5_conclusion.tex
\section{Conclusion}
In this work, we tackle the challenge of generating structured data using LLMs in real-world applications, where complex rules and natural language ambiguity often hinder the effectiveness of traditional methods. 
To address this, we introduce MLPrompt, a novel method that improves LLM reasoning to generate structured data by translating error-prone rules into another language, enhancing attention from LLM, and overall data quality. 
In comparison with state-of-the-art prompting strategies like CoT, ToT, and SC, MLPrompt demonstrates faster inference times and lower error rates. 
Additionally, we utilize MLPrompt to bridge the gap between LLM and autonomous industrial MIP generation, conducting extensive experiments on Text-to-MIP to prove MLPrompt’s effectiveness. 
Finally, we explore the possibility of applying MLPrompt to structured data generation tasks, such as Text-to-SQL.

%% file: Secs/6_limitation.tex
\section{Limitations}

\textbf{Difficulty in Identifying Rule Violations in Natural Language Prompts: }
While MLPrompt can enhance the quality of generated data after localizing the rule LLM would fail to follow, 
the identification of omitting or violated rules remains a significant challenge.
One limitation of our approach arises from the abstract nature of natural language, which often leads to rule inter-dependencies.
Unlike programming languages, which can be translated into executable code to verify rule compliance, natural language is an abstract representation that cannot be executed directly.
In structured data generation tasks like text-to-SQL (see Appendix \ref{sec:appendix:text2sql}, we define rules for LLMs to follow and validate the generated SQL by comparing it to the expected output. However, when discrepancies occur, pinpointing the specific rule violation is difficult, often requiring manual analysis.

However, enforcing rules with mathematical constraints, such as checking the data values, is relatively straightforward. 
The mechanism or model that can pinpoint which rule is violated is crucial for the effectiveness of MLPrompt.

\hfill \break
\textbf{Which non-dominant language should we use: }
In our experiment, the dominant language for LLMs is English, and we implement MLPrompt by translating the error-prone rule in English to another non-dominant language, such as German, French, Mandarin, Thai, Japanese, and Korean.
The performance of German and French is notably lower, while Mandarin, Thai, and Korean show better results.
Since our method relies on the GPT-4 series, and OpenAI has not disclosed the language distribution in GPT-4's training data, we reference the language distribution from GPT-3~\cite{brown2020language}. According to this, English, German, and French are among the top three languages in the training dataset, while Mandarin, Thai, and Korean rank much lower, around the 20s.

We hypothesize that MLPrompt is most effective when the distribution difference between dominant and non-dominant languages in the training data is neither too large nor too small. However, this hypothesis remains unproven, and selecting the appropriate language combination in the MLPrompt generation remains a key challenge.

%% file: Secs/7_appendix.tex
\appendix

\section{Appendix}
\label{sec:appendix}
\subsection{An Example for ComplexOR Dataset}
\label{sec:appendix:complexor}
Binpacking modeling problem in the ComplexOR dataset is shown blew:
\lstset{
    basicstyle=\ttfamily\fontsize{8}{10}, 
    breaklines=true,                       
    fontadjust=true,                       
    columns=fullflexible,                  
    keywordstyle=\bfseries,                
}

\begin{lstlisting}
{
  "id": 3,
  "title": "Binpacking Problem",
  "description": "The bin packing problem involves assigning items of known weights to bins with uniform capacity. The objective is to minimize the total number of bins utilized while ensuring that all items are allocated and each bin's total weight does not exceed the bin capacity.",
  "category": ["Binpacking Problem"],
  "model": {
    "set": [
      {
        "name": "I",
        "description": "Set of items"
      }
    ],
    "parameter": [
      {
        "name": "s",
        "description": "weight of item `i`",
        "domain": "{i <in> I}"
      },
      {
        "name": "c",
        "description": "Capacity of a bin"
      }
    ],
    "variable": [
      {
        "name": "y",
        "description": "Binary variable, 1 if we use bin `j`",
        "domain": "{j <in> I}",
        "type": "binary"
      },
      {
        "name": "x",
        "description": "Binary variable, 1 if we assign item `i` to bin `j`",
        "domain": "{i <in> I, j <in> I}",
        "type": "binary"
      }
    ],
    "objective": [
      {
        "name": "MinBins",
        "description": "Minimize the total number of used bins",
        "sense": "min",
        "function": "<sum>_{j <in> I} y_{j}"
      }
    ],
    "constraint": [
      {
        "name": "CapConstraint",
        "description": "The total weight of assigned items to a bin should not exceed the bin capacity",
        "domain": "{j <in> I}",
        "function": "<sum>_{i <in> I} s_{i} * x_{i,j} <= c * y_{j}"
      },
      {
        "name": "AssignConstraint",
        "description": "Every items should be assigned to a bin",
        "domain": "{i <in> I}",
        "function": "<sum>_{j <in> I} x_{i,j} = 1"
      }
    ]
  }
}
\end{lstlisting}

\subsection{Rule for Text-to-MIP}
\label{sec:appendix:text2mip}
The following rules are to be obeyed by the LLM, combined with modeling information (shown in Appendix \ref{sec:appendix:complexor} from the ComplexOR dataset), to generate the prompt.

\begin{lstlisting}

You must follow the following rules:

1: If the set in the model.json file has the 'range' key, this set has a hyper-parameter. For example, {range: [1,T]}, where T is the hyper-parameter. If the set in the model.json file doesn't have the 'range' key, the set doesn't have a hyper-parameter.
2: The number of set equals to the number of hyper-parameter.
3: In the set key, you need to append each set's lower bound and upper bound to the set value in order. The value of bound is generated by you after reading the model json file.
4: In the above case, if json_obj['model']['set'] does not have the range field, you need to add [null, null] to the hyper-parameter. If json_obj['model']['set']} has the range field, you need to add the predicted lower and upper bounds to the hyper-parameter.
5: The lower bound and upper bound of the set must be numbers, not null! The lb of set can start from non-one number.
6: When you provide the lower and upper bounds of the parameter, the value includes the lower bound but does not include the upper bound.
7: You should specify the data type of each parameter in order. The lb and ub of a parameter must either both be integers or both be float. If you think the type of this parameter is int, then you should append 'integer'. If you think the type of this parameter is float, then you should append 'float'.
8: You should ensure that the gap between ub and lb of the parameter should be less than or equal to 15. ub-lb <= 15! At the same time, diversify the upper and lower bound of parameter.
9: You must follow the json data format {'set': [[lb1, ub1], [lb2,ub2]...], 'hyper-parameter': [[lb1, ub1], [lb2,ub2]...], 'parameter': [[lb1, ub1],[lb2, ub2]...], 'parameter_types':[integer, integer, float, ...]}.
\end{lstlisting}

\begin{figure}[H]
\centering
\includegraphics[width=0.5\textwidth]{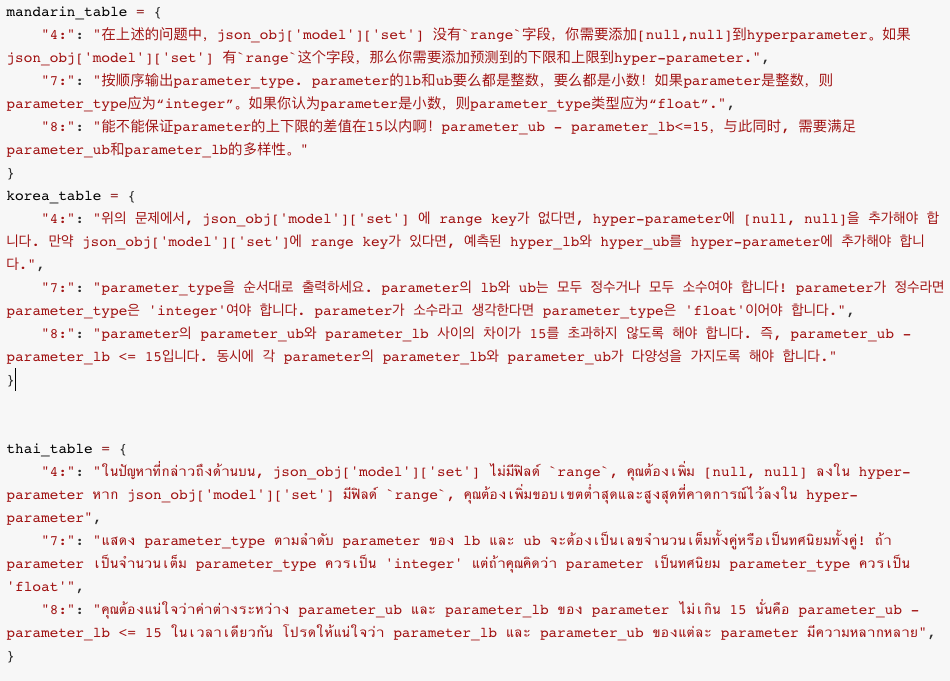}
\caption{The figure shows the translations of rules 4, 7, and 8 into Mandarin, Korean, and Thai, illustrating the language-specific versions of these rules.
}
\label{fig:language_table}
\end{figure}
Since our experiment primarily focuses on generating data according to rules 4, 7, and 8, we also present the corresponding dictionary in Fig.~\ref{fig:language_table}, which illustrates how these rules are translated into Mandarin, Korean, and Thai.

\subsection{Prompt Template for Text-to-MIP}
\label{sec:appendix:text2mip_prompt}
The MLPrompt approach involves simply replacing a specific rule with its equivalent in another language.
The dominant language in the prompt is English, and during application, only the target rule needs to be swapped with its counterpart in the desired language. The prompt template is as follows:

\begin{tcolorbox}[colback=yellow!5!white, colframe=black!75!black, title=Prompt Template, breakable]
You are required to return a feasible solution distribution under the given constraints.

Please read the following mixed integer programming (MIP) model and return a JSON object containing the lower and upper bounds for each set, hyper-parameter, and parameter.

Since the model does not include any data, your task is to provide the lower and upper bounds for every set, parameter, and hyper-parameter to construct instances for this model.

The required JSON format is as follows: 

\{'set': [[lb1, ub1], [lb2,ub2]...], 'hyper-parameter': [[lb1, ub1], [lb2,ub2]...], 'parameter': [[lb1, ub1],[lb2, ub2]...], 'parameter\_types':[integer, integer, float, ...]\}.

\textcolor{red}{\textbf{Rules} is shown in Appendix \ref{sec:appendix:text2mip}. }

\textcolor{blue}{\textbf{Modeling Information} from ComplexOR Dataset. (An example is shown in Appendix \ref{sec:appendix:complexor})}
\end{tcolorbox}

\subsection{Analysis rules for Text to MIP}
\label{sec:appendix:analysis_text2MIP}
\textbf{Rule 4} requires the LLM to read through the detailed modeling information, which consists of a long text, analyze the modeling process, and determine whether each set contains a corresponding hyper-parameter after reading the rule. Rule 4 evaluates the LLM's ability to comprehend the lengthy text and reason the result based on both the given question and the context.

\textbf{Rule 7} requires consistency between the generated parameter values and their respective data types. Since the LLM must first output all parameter values and then generate their corresponding data types, it often forgets previous outputs, leading to struggles in maintaining coherence across the generated values. Rule 7 assesses the LLM's ability to ensure coherence throughout the process.

\textbf{Rule 8} requires the LLM to diversify its output while adhering to specific mathematical constraints. The LLM often fails by generating common or repeated data values and not complying with the necessary constraints. Rule 8 evaluates the LLM's ability to generate diverse numbers, perform calculations, and adhere to given constraints.

\subsection{Details of CoT, ToT, and SC}
\label{sec:appendix:SOTA_prompt_methods}
Due to the inherent ambiguity in natural language, evaluating intermediate steps during the reasoning process becomes problematic for methods like CoT, ToT, and SC. 
These methods rely on breaking down complex tasks into smaller steps and verifying them, which works well in structured environments like logic expressions. 
However, in structured data generation tasks, it is challenging to clearly define whether an intermediate result is accurate, as language is often subject to interpretation.
Additionally, the interconnected nature of rules in structured outputs, such as JSON, makes it difficult to verify each step in isolation—correctness in one part does not necessarily guarantee overall correctness when the parts are combined.
In our experiment, we made slight modifications to CoT, ToT, and SC to adapt them to our specific task. The details of these modifications are as follows:

\textbf{CoT} We implement CoT using zero-shot prompting by simply adding the phrase, "Let's think it step by step," to guide the LLM in reasoning through the task.

\textbf{ToT} We implement ToT by having the LLM generate each part of the desired JSON, similar to the structure of a tree, one by one. The previous output is combined into the prompt at each step until the entire requirement is fulfilled.

\textbf{SC} We implement SC by generating the full JSON multiple times, evaluating each result as either correct or incorrect. These labeled outputs are then combined into a new prompt, which is provided to the LLM to generate a more consistent and accurate final output.

\subsection{Text to SQL}
\label{sec:appendix:text2sql}

To demonstrate the generalization and robustness of MLPrompt, we apply our method to the Text-to-SQL task using the Spider V1.0 dataset \cite{yu2018spider}. 
Given that each entity—such as companies, individuals, and datasets—follows its own coding style, we filter questions that GPT-4 answered incorrectly by having it generate SQL queries directly based on the SQL schema used in database construction, without any specific rules. 
We manually analyzed these incorrect cases and summarized a set of rules for GPT-4 to follow. 
Given the large size of the dataset, we used only 20\% of it and identified the types of questions GPT-4 struggles to handle effectively. 
By combining these rules with the SQL schema, we employed zero-shot learning with GPT-4 to test whether it could generate the correct queries. The rules are as follows:

\begin{tcolorbox}[colback=yellow!5!white, colframe=black!75!black, title=Rules for Text-to-SQL, breakable]
     1. When handling queries involving counting-related issues, avoid using "LEFT JOIN" or "RIGHT JOIN" to generate non-existent records. Instead, use "INNER JOIN".
     
     2. Pay attention to the order of values requested in the question! Make sure the "SELECT" clause provides the appropriate field order. If the question does not ask for a count, do not include it. If the requirement is "xxx1 for each xxx2," the first field in "SELECT" should contain the data (xxx1) and the last field should return the category (xxx2). If the target is a primary key and the question asks how many xxx each primary key has, put the primary key last and the count first. There's no need to include non-key fields like names.
     
    3. If the logic is complex, use subqueries instead of joining tables. Avoid using "DISTINCT" unless necessary. When considering relationships between tables based on the query and the table’s information, determine whether it's a "has-a" or "is-a" relationship. If the "SELECT" fields include a "has-a" scenario, use "DISTINCT" to avoid duplicates. If the query asks to list all xxx, do not use "DISTINCT".
    
     4. If the user wants to find the maximum/minimum/average value in a table, consider using subqueries instead of grouping by different categories. If the query requires finding the maximum/minimum/average value within categories, use aggregation functions and "GROUP BY".
     
     5. If the query requires listing all information, use "SELECT *".
\end{tcolorbox}


